\documentclass[runningheads]{llncs}

\usepackage{eccv}

\usepackage{eccvabbrv}

\usepackage{graphicx}
\usepackage{booktabs}
\usepackage{multirow}
\usepackage{makecell}
\usepackage{subcaption}
\usepackage[accsupp]{axessibility}  

\usepackage{hyperref}

\usepackage{orcidlink}

\begin{document}

\title{HNDiff: Haze-Noise Diffusion for Image Dehazing} 

\titlerunning{HNDiff}

\author{Jin-Ting He\inst{1}\orcidlink{0009-0002-5414-0693} \and
Fu-Jen Tsai\inst{2}\orcidlink{0000-0003-1486-2229} \and
Yan-Tsung Peng\inst{3}\orcidlink{0000-0002-3802-1670}
\and
Min-Hung Chen\inst{4}\orcidlink{0000-0002-4046-3937}
\and
Chia-Wen Lin\inst{2}\orcidlink{0000-0002-9097-2318}
\and
Yen-Yu Lin\inst{1}\orcidlink{0000-0002-7183-6070}}

\authorrunning{J.-T.~He et al.}

\institute{Department of Computer Science, \\
National Yang Ming Chiao Tung University, Taiwan\\
\email{jinting.cs12@nycu.edu.tw,  lin@cs.nycu.edu.tw}
\and
National Tsing Hua University, Taiwan\\
\email{fjtsai@gapp.nthu.edu.tw, cwlin@ee.nthu.edu.tw}
\and
National Chengchi University, Taiwan\\
\email{ytpeng@cs.nccu.edu.tw}
\and
NVIDIA, Taiwan\\
\email{minhungc@nvidia.com}
}
\maketitle

\begin{abstract}

Existing diffusion-based methods have recently made significant progress in image dehazing. However, they typically neglect the physics of haze formation and reconstruct clean images from pure Gaussian noise, thereby limiting their restoration potential. 
To address this issue, we propose Haze-Noise Diffusion (HNDiff), a novel diffusion framework that embeds the atmospheric scattering model as an inductive bias. By grounding diffusion in physical principles, HNDiff ensures that the restoration aligns more closely with underlying mechanisms of haze formation.
In its forward process, we introduce joint haze-noise diffusion with a haze-aware noise scheduler, which progressively adds both haze and noise to an image.
Essentially, the scheduler adapts noise levels according to haze density, meaning that regions with heavier haze receive stronger noise injection to encourage content generation, while clearer regions receive lighter noise to better preserve details, which directly links the forward degradation process with the physics of haze.
In the reverse process, we then derive a physically consistent dehazing-denoising process that simultaneously removes haze and noise to restore a clean image in a manner aligned with the forward degradation process.
To further enhance practicality, we propose Latent HNDiff, which compiles clean latent priors that can be seamlessly integrated into existing dehazing networks to boost performance. 
Extensive experiments show that our work significantly improves leading dehazing backbones and achieves state-of-the-art results on benchmark datasets.
The project page is available at \url{https://jin-ting-he.github.io/HNDiff/}.

%
%
%
%
%
\end{abstract}
\section{Introduction}
Hazy weather conditions caused by atmospheric scattering frequently degrade image visibility by reducing contrast and obscuring scene details. 
Such degradation not only impairs human perception but also severely hinders the performance of many vision applications, such as object detection~\cite{Kim_2024_CVPR, Wang_2024_CVPR}, semantic segmentation~\cite{Benigmim_2024_CVPR, Weber_2024_CVPR}, and face recognition~\cite{Minchul_2024_CVPR, Mi_2024_CVPR}.
To address the challenges, single image dehazing has emerged as a feasible solution to restore a clear image from a single hazy input. However, such a task remains highly ill-posed due to the complex interplay of scattering coefficients, atmospheric light, and scene depth.

%

\begin{figure*}[t!]
  \centering
  \includegraphics[width=1.0\textwidth]{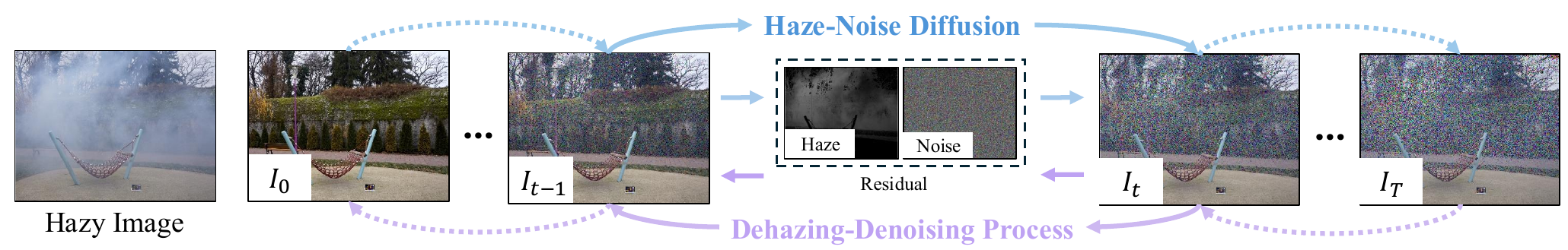} %
  \caption{
   HNDiff leverages the ASM inductive bias, progressively adding haze and noise in the forward process and removing them in the reverse process for image dehazing.
  }
  \label{fig:teaser}
\end{figure*}

Driven by advances in deep learning, CNN-based methods~\cite{dong2020multi, wu2021contrastive, bai2022self, cui2023focal} have achieved impressive results in image dehazing. 
Transformer-based approaches~\cite{qin2020ffa, guo2022image, qiu2023mb, cui2024revitalizing, fang2025guided} further improved performance by exploring long-range dependencies and global context.
Recently, Mamba-based methods~\cite{zheng2024u, li2025mair} have emerged as an efficient alternative with linear computational complexity.
Despite the advancements, these methods still struggle in heavy haze scenarios, where most information is lost, leading to limited restoration quality.

%

In parallel, diffusion models~\cite{ho2020denoising, rombach2022high} have shown strong generative ability in image synthesis, producing results with rich details and sharp textures. 
Motivated by this progress, several studies~\cite{yang2024unleashing, wang2025learning} have applied diffusion algorithms to image dehazing.  
Yet, conventional diffusion models are fundamentally misaligned with the nature of haze. That is, they reconstruct clean images from pure Gaussian noise, and their stochastic nature~\cite{ye2024learning} often causes deviations from the original image, thereby reducing restoration fidelity.
More importantly, they usually neglect the physical properties of haze formation, resulting in suboptimal restoration performance.
%
%
%
According to the Atmospheric Scattering Model (ASM)~\cite{narasimhan2003contrast}, a hazy image results from attenuated scene radiance and global atmospheric light. 
Haze density may vary spatially across the image, as it depends on both the light scattering effect and scene depth, and intensifies with the increasing scattering coefficient. 
Thus, haze exhibits structured, spatially varying degradations, unlike the random Gaussian noise typically assumed in conventional diffusion models.

Based on this observation, we present {\em Haze-Noise Diffusion} (HNDiff), a new framework that redefines the forward process through a haze-noise diffusion mechanism. Instead of injecting only Gaussian noise, we integrate the ASM into the diffusion process, mimicking the physical formation of haze by highlighting its spatially varying characteristics.
%

In the forward process, HNDiff carries out the haze-noise diffusion mechanism, which gradually adds both Gaussian noise and haze to a clean image, as illustrated in~\cref{fig:teaser}. 
To better control this process, we introduce the haze-aware noise scheduler, which dynamically adjusts the noise level according to haze density: hazier regions are assigned higher noise to boost generative capacity, while clearer regions receive less noise to preserve detail fidelity.  
Progressive haze diffusion and adaptive noise scheduling require transmission maps from ASM, which are generally unavailable. 
To overcome this limitation, we consider the haze residual, defined as the incremental haze accumulated as the scattering coefficient increases. We develop a continuous accumulation formulation to represent this residual implicitly in HNDiff and thus eliminate the need for explicit transmission maps. 
Through this design, the forward process remains ingeniously consistent with ASM, enabling progressive haze addition in tandem with adaptive noise injection.

%
%

In the reverse process, we derive the dehazing-denoising process, which is grounded by the physical principles of ASM and can implicitly approximate noise and haze residuals through dedicated estimators, thereby removing haze and noise to restore clean images. 
However, directly applying diffusion in the image space incurs substantial computational overhead and may suffer from fidelity issues in severely degraded regions due to the stochastic nature of the diffusion process. 
To address these problems, we propose Latent HNDiff, a prior generation network that integrates flexibly with dehazing backbones, allowing for more accurate and visually consistent restoration. This latent approach not only reduces computational cost but also enhances the applicability of the framework across diverse dehazing models.

%


The key contributions of our work are summarized as follows:
\begin{itemize}
  \item We propose HNDiff, a novel diffusion-based framework that incorporates ASM as an inductive bias, specifically designed for image dehazing.
  \item HNDiff implements a haze-noise diffusion forward process that adds both haze and noise, and a corresponding dehazing-denoising reverse process with two dedicated estimators to respectively remove haze and noise.
  \item We design the haze-aware noise scheduler to adaptively adjust noise levels based on haze densities.
  \item Extensive experiments demonstrate that HNDiff consistently improves four representative dehazing models and achieves state-of-the-art performance on seven benchmark datasets.
\end{itemize}


\section{Related work}
\subsection{Image Dehazing}
\textbf{CNN-based Dehazing.} Deep learning has revolutionized image dehazing with CNN-based methods~\cite{dong2020multi, wu2021contrastive, bai2022self, yang2022self, cui2023focal, tsai2025phatnet} achieving impressive breakthroughs. 
For instance, Dong~\etal~\cite{dong2020multi} proposed a boosted decoder combined with a dense feature fusion module to progressively restore images. 
Wu~\etal\cite{wu2021contrastive} introduced contrastive regularization within an autoencoder for efficient dehazing.
More recently, Cui~\etal\cite{cui2023focal} presented a dual-domain selection mechanism and an efficient multi-scale network to further enhance restoration quality.
%
%
%
%
%
%

\noindent\textbf{Transformer-based Dehazing.}
In addition to CNNs, Transformer-based methods have shown great promise in image dehazing by leveraging attention mechanisms to model long-range dependencies and global context~\cite{qin2020ffa, guo2022image, song2023vision, qiu2023mb, valanarasu2022transweather,  cui2024revitalizing, fang2025guided}.
For example, Qiu~\etal~\cite{qiu2023mb} approximated softmax-attention with a Taylor expansion to achieve linear complexity for effective dehazing. 
Cui~\etal~\cite{cui2024revitalizing} designed a multi-shape attention module with rectangle and dilated operations to enlarge receptive fields. 
Fang~\etal~\cite{fang2025guided} integrated phase and attention modules to leverage YCbCr textures for recovering clearer features.
%
%
%
%
%
%

\noindent\textbf{Mamba-based Dehazing.} Mamba-based methods have recently emerged as efficient alternatives for image dehazing, capturing global context with linear computational complexity. 
Zheng \& Wu\cite{zheng2024u} combined convolution for local feature extraction with state space models to capture long-range dependencies in dehazing. 
Li~\etal~\cite{li2025mair} designed an S-shaped stripe-based scanning strategy to better preserve locality and continuity for more effective restoration.

\noindent\textbf{ASM-based Dehazing.} Beyond architectural advances, several studies~\cite{shao2020domain, chen2021psd, yang2022self, wu2023ridcp, fang2024real, shin2025hazeflow, Huang_2025_BMVC} explicitly exploit the Atmospheric Scattering Model (ASM) to improve dehazing.
Wu~\etal~\cite{wu2023ridcp} designed an ASM-based data generation pipeline to synthesize hazy images for network training. 
Fang~\etal~\cite{fang2024real} derived a cooperative unfolding network directly from ASM, jointly optimizing the transmission map and the clean image. 
%
%

Despite these advancements, most dehazing methods are still trained end-to-end as direct regressors from hazy inputs to clean outputs. Although some ASM-based approaches incorporate the ASM as physical guidance to constrain this mapping, both regression-based and ASM-based methods often struggle under extremely dense haze, as the severe information loss makes it difficult to recover realistic high-frequency details. In contrast, our method couples the ASM with a diffusion process and leverages the generative capability of noise diffusion to compensate for missing content and restore plausible fine structures in heavily degraded regions.


\subsection{Diffusion Models}
\textbf{Diffusion for Low-level Vision.}
Diffusion models~\cite{ho2020denoising, rombach2022high, mengsdedit, zhang2023adding, wu2024id} have shown strong generative capability in image synthesis and can produce results with rich details and sharp textures through forward noise diffusion and reverse denoising. 
This success has inspired much research exploring their potential in diffusion algorithms for various low-level vision tasks~\cite{zhang2024diffusion, garber2024image, li2024rethinking, liu2024residual, liu2024diff, xia2023diffir, Zheng_2024_CVPR, rajagopalan2025gendeg, luo2025visual, hereti, he2024domain, rao2024rethinking}. 
For example, 
He~\etal~\cite{he2024domain} introduced a domain-adaptive blur condition to guide a diffusion-based blurring model, enabling the generation of domain-adaptive blurred videos.
Xia~\etal~\cite{xia2023diffir} employed diffusion models to extract compact priors used to guide a dynamic transformer for image recovery.
Liu~\etal~\cite{liu2024diff} utilized a pre-trained diffusion model with task-specific priors for diverse image restoration tasks.
Luo~\etal~\cite{luo2025visual} presented visual instruction-guided diffusion that models degradation patterns for all-in-one image restoration.
%
%
%
%
%

\noindent\textbf{Diffusion for Image Dehazing.}
Within low-level vision, several studies have focused specifically on image dehazing using diffusion models~\cite{yang2024unleashing, wang2024frequency, liu2024diff, wang2025learning, liu2025frequency}.
For instance, Yang~\etal~\cite{yang2024unleashing} exploited the semantic latent space of a pre-trained diffusion model to guide dehazing without retraining.
Wang~\etal~\cite{wang2025learning} combined diffusion-based hazy image generation with accelerated fidelity-preserving sampling for efficient dehazing.
Liu~\etal~\cite{liu2025frequency} leveraged diffusion models in the frequency domain with an amplitude residual encoder to enhance unpaired image dehazing. 
Despite these advances, these methods rely on conventional noise diffusion initialized with pure Gaussian noise, which disregards the physical properties of haze formation. 
As a result, the stochastic nature~\cite{ye2024learning} of the process often leads to deviations from the target restoration fidelity, resulting in degraded performance and less consistent visual quality.
%
%
%
%
%
%
%

\noindent\textbf{Degradation-aware Diffusion.}
Recently, a few studies~\cite{liu2024residual, zhou2025physics, he2025blurdm} have explored degradation-aware diffusion for image restoration. 
For example, Liu~\etal~\cite{liu2024residual} proposed residual diffusion and operated on the difference between hazy and clean images. 
He~\etal~\cite{he2025blurdm} designed blur diffusion that implicitly models the blur formation process through a dual-diffusion forward scheme for image deblurring. Zhou~\etal~\cite{zhou2025physics} introduced a physics-guided dehazing diffusion by reformulating haze accumulation as a time-indexed process. 
However, these approaches either neglect the role of physical scene transmission in modeling haze degradation within the diffusion process or overlook haze density in the noise scheduling.
To address these issues, our approach embeds ASM into the diffusion process as an inductive bias and adaptively adjusts noise addition according to haze density, achieving significantly improved dehazing performance.

%

%

\section{Method}

\begin{figure*}[t!]
  \centering
  \includegraphics[width=0.98\textwidth]{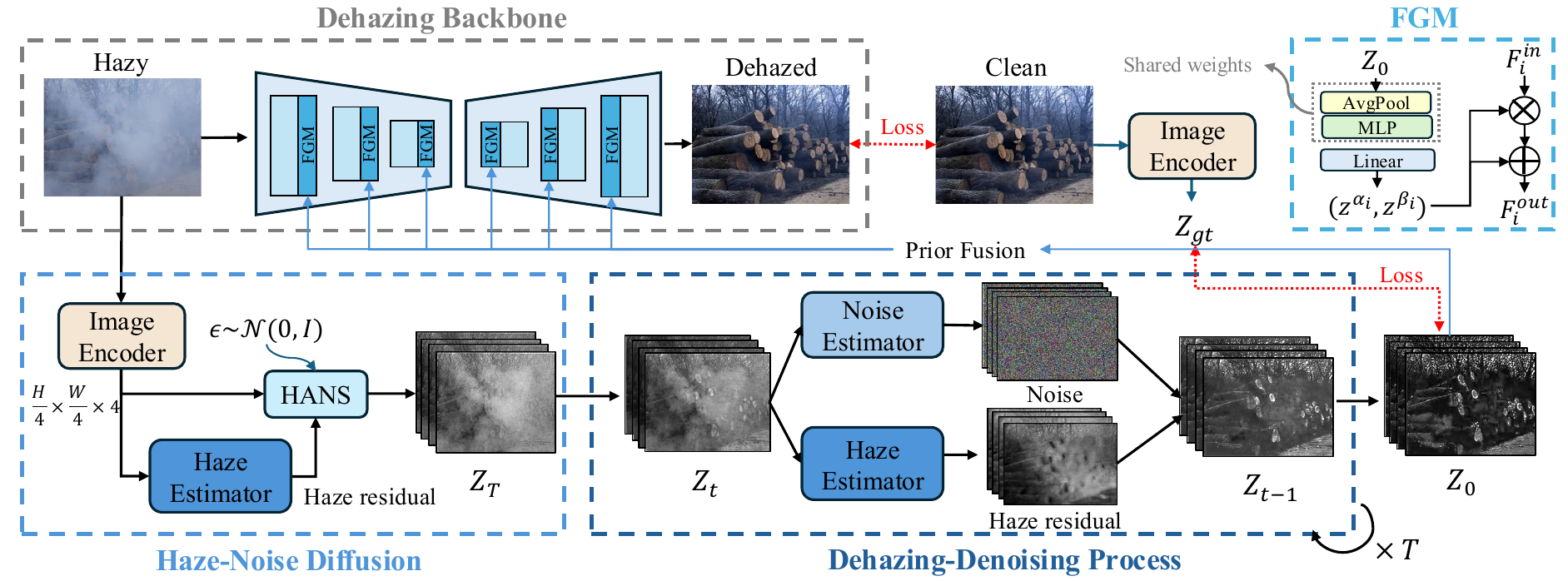} %
  \caption{
  \textbf{Overview of Latent HNDiff.} 
  The framework starts by using an image encoder to extract latent priors from a hazy input. 
  These priors undergo a haze-noise diffusion process to produce the diffused hazy and noisy representation $Z_T$. 
  During the reverse process, dehazing and denoising are performed jointly by iteratively estimating both noise and haze residuals to recover clean priors $Z_0$. Lastly, the recovered priors are integrated into a dehazing backbone via the Feature Gating Module (FGM) to improve restoration quality.
  }
  \label{fig:pipeline}
\end{figure*}

This section presents the proposed {\em Haze-Noise Diffusion} (HNDiff), a novel framework that integrates the Atmospheric Scattering Model (ASM)~\cite{narasimhan2003contrast} into the diffusion process for image dehazing.
As depicted in~\cref{fig:teaser}, HNDiff defines a physics-guided forward {\em Haze-Noise Diffusion} process equipped with a {\em Haze-Aware Noise Scheduler} (HANS) and a reverse {\em Dehazing–Denoising Process}.
In the forward direction, an input image is progressively degraded by jointly introducing haze and Gaussian noise as the scattering coefficient increases, while HANS adaptively controls the noise level based on the local haze density, ensuring the corruption remains consistent with ASM-guided haze formation.
In the reverse direction, HNDiff starts from a hazy input perturbed by Gaussian noise and iteratively removes both haze and noise in a manner consistent with the ASM.
For high-fidelity restoration and better efficiency, as illustrated in~\cref{fig:pipeline}, we further propose {\em Latent HNDiff}, where HNDiff serves as a prior generation network in the latent space, and the learned prior is injected into a dehazing backbone via a Feature Gating Module (FGM), enabling plug-and-play enhancement of existing dehazing architectures. 
In the following, \cref{sec:haze-noise-diffusion} details the Haze-Noise Diffusion process and HANS, \cref{sec:dehazing-denoising-process} presents the Dehazing–Denoising Process, and \cref{sec:latent-hndiff} describes Latent HNDiff together with the FGM integration.

\subsection{Haze-Noise Diffusion}
\label{sec:haze-noise-diffusion}

Haze in image formation arises from atmospheric scattering, where scene radiance is attenuated during transmission and blended with global atmospheric light, leading to reduced visibility and a loss of detail.
This process can be mathematically modeled by the ASM:  
\begin{equation}
I_H(x) = I_{0}(x)\, \tau(x) + A \,(1 - \tau(x)), \mbox{ where }\tau(x) = e^{-\sigma(x) d(x)}
\label{eq:ASM}
\end{equation}
with $x$ denoting the pixel index.
In \cref{eq:ASM}, $I_H\in \mathbb{R}^{H \times W \times3}$, $I_{0} \in \mathbb{R}^{H \times W \times3}$, $\tau\in \mathbb{R}^{H \times W \times1}$, $A\in \mathbb{R}^{3}$, $\sigma \in \mathbb{R}^{H \times W \times1}$, and $d\in \mathbb{R}^{H \times W \times1}$ denote the hazy image, the clean scene radiance, the transmission map, the global atmospheric light, the scattering coefficient, and the scene depth, respectively. 
This formulation explicitly models haze formation, where larger scattering coefficients $\sigma$ or greater depths $d$ yield smaller transmission $\tau$, increasing haze density and reducing scene visibility.  
However, existing diffusion-based dehazing methods typically adopt conventional diffusion models that add zero-mean Gaussian noise and drive the image toward pure noise, which does not reflect the structured, spatially varying haze described in \cref{eq:ASM}. To bridge this gap, we introduce a \emph{haze–noise diffusion} process in which ASM-based haze formation acts as a mean shift of the Gaussian corruption, so that the forward process follows a physically meaningful clean-to-hazy evolution while stochastic noise is injected around this trajectory.

\noindent\textbf{Forward Haze-Noise Diffusion.}  
In the forward process of HNDiff, we propose a haze-noise diffusion that embeds the physical haze formation process into noise diffusion. 
Specifically, a clean image is progressively degraded by both haze and Gaussian noise. 
The forward transition at time step $t$ is defined as
\begin{equation}
I_t(x) = I_{t-1}(x) e^{-\alpha_t \sigma(x) d(x)} + A \left( 1 - e^{-\alpha_t \sigma(x) d(x)} \right) + \beta_t(x) \epsilon_t(x),
\label{eq:forward1}
\end{equation}
where $I_t$ denotes the intermediate hazy and noisy image at step $t$, $\alpha_t$ is the scaling factor, $\epsilon_t \sim \mathcal{N}(0, \mathbf{I})$ represents Gaussian noise, and $\beta_t(x)$ is the noise scaling coefficient at pixel $x$. 
As seen in \cref{eq:forward1}, $I_{t-1}$ is further degraded according to \cref{eq:ASM}, with scene radiance attenuated by the $\alpha_t$-scaled scattering coefficient, while noise is injected with another scaling coefficient $\beta_t(x)$.

\noindent\textbf{Haze-Aware Noise Scheduler.}  
\label{section:hans}
Since haze density varies spatially, the generative capacity controlled by noise diffusion should also adapt across pixels.
We therefore introduce a haze-aware noise scheduler, which defines the pixel-wise noise scaling coefficient as
$
\beta_t(x) = 1 - e^{-\alpha_t \sigma(x) d(x)}.
$ 
This design makes the injected noise explicitly dependent on haze density: pixels with heavier haze receive larger $\beta_t(x)$, thus introducing stronger noise that triggers diffusion to reconstruct severely degraded details; conversely, pixels with lighter haze yield smaller $\beta_t(x)$, injecting less noise to preserve content fidelity through regression. 
This adaptive scheduling enables the forward process to jointly model haze degradation and stochastic corruption.

\noindent\textbf{Sampling Probability and Reparameterization.}  
Inspired by~\cite{liu2024residual, he2025blurdm}, we regard degradation in the forward process as a deterministic mean shift. From \cref{eq:forward1}, each step from $I_{t-1}$ to $I_t$ can thus be expressed as a Gaussian transition, where the mean is shifted by haze and stochastic perturbations are introduced by Gaussian noise:
\begin{align}
&q(I_t(x) \mid I_{t-1}(x), \phi) 
\\&:= \mathcal{N}\!\left(I_t(x) \;\middle|\; I_{t-1}(x) e^{-\alpha_t \sigma(x) d(x)} + A \left( 1 - e^{-\alpha_t \sigma(x) d(x)} \right), \, \beta_t^2(x) \right),
\label{eq:forward2}
\end{align}
where $\phi = \{d(x), \sigma(x), A\}$. By iterating \cref{eq:forward2}, we obtain a sequence of progressively hazy and noisy images $\{ I_{1}, I_{2}, \ldots, I_{T} \}$ through a $T$-step diffusion process, with the complete forward sampling probability $q(I_{1:T}(x) \mid I_{0}(x), \phi) = \prod_{t=1}^{T} q(I_t(x) \mid I_{t-1}(x),\phi).$  
%
However, existing dehazing datasets provide only hazy-clean image pairs and do not include $\phi$ (i.e., atmospheric light, scattering coefficients, and scene depth necessary to compute transmission). 

To address this limitation, we apply the reparameterization trick~\cite{ho2020denoising} to \cref{eq:forward2} and obtain the conditional distribution after $T$ steps as  
\begin{align}
&q(I_{T}(x) \mid I_{0}(x), \phi) \\&= \mathcal{N}\!\left(I_T(x) \;\middle|\; I_{0}(x) e^{-\sum_{t=1}^T \alpha_t \sigma(x) d(x)} + A \left( 1 - e^{-\sum_{t=1}^T \alpha_t \sigma(x) d(x)} \right), \, \!\bar{\beta}_T^2(x) \right),
\label{eq:forward4}
\end{align}
where $\alpha_t = \tfrac{1}{T}, \; \forall t \in \{1,2,\ldots,T\}$, and  
$
\bar{\beta}_T(x) = \sqrt{\frac{(1 - e^{-(1/T) \sigma(x) d(x)})(1 - e^{-2\sigma(x) d(x)})}{1 + e^{-(1/T) \sigma(x) d(x)}}}.
$
The complete derivation of \cref{eq:forward4} is provided in the supplementary materials (Sec.~1).
It follows that the hazy and noisy image $I_T$ can be sampled from $q(I_{T} \mid I_{0})$ via  
\begin{equation}
I_T(x) = I_{0}(x) e^{-\sigma(x) d(x)} + A \left( 1 - e^{-\sigma(x) d(x)} \right) + \bar{\beta}_T(x) \epsilon(x) = I_H(x) + \bar{\beta}_T(x) \epsilon(x),
\label{eq:forward5}
\end{equation}
where $I_T$ is generated in a single step by injecting noise into the hazy image $I_H$ via the haze-aware noise scheduler. This formulation preserves the Gaussian nature of the diffusion process while embedding ASM directly into the mean of the distribution through a physically grounded shift. As $\bar{\beta}_T(x)$ still relies on the transmission map, we introduce a learnable haze estimator to implicitly approximate it. The optimization details for the haze estimator are provided in~\cref{sec:latent-hndiff}.
\subsection{Dehazing-Denoising Process}
\label{sec:dehazing-denoising-process}
In the reverse generation procedure, we aim to progressively remove both haze and noise from the degraded observation $I_T$ to recover the clean image $I_0$. 
Unlike conventional diffusion models that start from pure Gaussian noise, our method initializes from the hazy-noisy sample $I_T$ drawn from the Gaussian distribution \cref{eq:forward4}.
Inspired by the deterministic sampling formulation in \cite{songdenoising}, we define the reverse transition distribution as
\begin{equation}
    p_\theta(I_{t-1}(x) \mid I_t(x)) = q_\delta(I_{t-1}(x) \mid I_t(x), I_0(x), \phi). 
    \label{eq:transition_prob}
\end{equation}
The transition probability $q_\delta$ in \cref{eq:transition_prob} is defined as
\begin{align}   
& q_\delta(I_{t-1}(x) \mid I_t(x), I_0(x), \phi)  = \mathcal{N} \left( I_{t-1}(x) \mid \mu_t(x), \delta^2_t(x) \right), \label{eq:reverse1} \quad \text{where} \\
    & \mu_t(x) = I_0(x) e^{-\sum_{s=1}^{t-1} \alpha_s \sigma(x) d(x)} + A \left( 1 - e^{-\sum_{s=1}^{t-1} \alpha_s \sigma(x) d(x)} \right) \\& \quad \quad \quad+ \sqrt{\bar{\beta}_{t-1}^2(x)-\delta_t^2(x)}\epsilon_{t-1}(x),
\end{align}
and $\delta_t^2 = \eta \cdot \tfrac{\beta_t^2 \bar{\beta}_{t-1}^2}{\bar{\beta}_t^2}$ is a variance term that controls sampling stochasticity. 
When $\eta=0$, this yields a deterministic sampling. From \cref{eq:forward4}, we can derive 
\begin{equation}
I_0(x) = (I_t-(1-e^{-\sum_{s=1}^{t-1} \alpha_s \sigma(x) d(x)})A-\bar{\beta}_t\epsilon_t)  e^{\sum_{s=1}^{t-1} \alpha_s \sigma(x) d(x)}.
\label{eq:I_0}
\end{equation}
By substituting \cref{eq:I_0} into \cref{eq:reverse1} and simplifying, we obtain the sampling equation for $I_{t-1}(x)$ as
\begin{align}
    I_{t-1}(x) = \left( I_t(x) - N_t(x) \left( 1 - e^{-\alpha_t\sigma(x) d(x)}  \right) \right) e^{\alpha_t\sigma(x) d(x)},
    \label{eq:reverse3}
\end{align}
where $N_t(x) = A + \epsilon_t(x)$ denotes the atmospheric noise, which is composed of the atmospheric light term and a Gaussian noise term. 
To reconstruct $I_0$, we iterate \cref{eq:reverse3} with two learnable estimators.
One is the noise estimator $N_t^\theta(I_t, I_H, t)$, which approximates $N_t$. 
The other is the haze estimator $1 - e^{-\alpha_t o^\theta(I_t, I_H, t)}$, which approximates the residual transmission term $1 - e^{-\alpha_t \sigma d}$ (the complement of the transmission), where $o^\theta(I_t, I_H, t)$ is a learnable network estimating the scattering–depth product $\sigma d$. 
%
%
Complete derivations of the variational lower bound and the sampling formulation are provided in the supplementary materials (Secs.~2 and~3).
In the following, we detail the optimization of the haze estimator and noise estimators in the latent space.

\subsection{Latent \texorpdfstring{HNDiff}{D2Diff}}
\label{sec:latent-hndiff}
Performing diffusion-based restoration directly in image space, as noted in \cite{rombach2022high, chen2023hierarchical, he2025blurdm}, can be computationally expensive and often result in slower, less stable optimization, with limited reconstruction fidelity. 
To address these challenges, and inspired by previous works \cite{rombach2022high, chen2023hierarchical, xia2023diffir, he2025blurdm}, we present Latent HNDiff.
%
As illustrated in~\cref{fig:pipeline}, our method runs the haze-noise diffusion procedure on an encoded representation extracted from the input image to generate informative priors, which are then injected into a dehazing backbone via the proposed Feature Gating Module (FGM).
By incorporating physically grounded haze formation into the diffusion procedure, we encourage the learned priors to capture haze-aware cues that ultimately improve restoration quality.

\noindent\textbf{Training strategy.}
Following the training scheme from~\cite{he2025blurdm}, we first pretrain a dehazing backbone equipped with an Image Encoder (IE) and a Feature Gating Module (FGM) using paired data $(I_H,I_0)$.
The ground-truth prior is extracted as
$
Z_{gt}=\mathrm{IE}\!\left(\mathrm{Concat}(I_H,I_0)\right),
$
and fused into multi-scale features $F_i^{in}$ through FGM:
\begin{align}
z_{gt}&=\mathrm{MLP}\!\left(\mathrm{AvgPool2D}\!\left(\mathrm{Unshuffle}(Z_{gt})\right)\right) \in \mathbb{R}^{1 \times C}; \label{eq:11}\\
F_i^{out}&=F_i^{in}\times z^{\alpha_i}+z^{\beta_i},\ \ \mbox{where } (z^{\alpha_i},z^{\beta_i})=\mathrm{Linear}(z_{gt}). \label{eq:12}
\end{align}
Next, we use the proposed HNDiff framework to learn a prior from the degraded representation $Z_T$ (obtained using the haze-noise diffusion process) and then recover the clean priors $Z_0$, supervised by
$
\mathcal{L}_{\text{prior}}=\lVert Z_0-Z_{gt}\rVert_1.
$
Lastly, we jointly fine-tune the IE, HNDiff, FGM, and the dehazing backbone, using the learned $Z_0$ to reconstruct the dehazed image $I_{dehz}$, supervised by $I_0$ under the standard backbone losses. More details are provided in supplementary materials (Sec. 4).

\section{Experiments}
\subsection{Experimental Setup}
\label{section:4.1}
\noindent\textbf{Implementation Details.}
HNDiff is composed of four key components: the Image Encoder (IE), the Feature Gating Module (FGM), the Haze Estimator, and the Noise Estimator. 
The IE consists of six residual blocks and four CNN layers, while the FGM is implemented with a pooling operation and a lightweight MLP. 
Both the Haze Estimator and Noise Estimator share the same network architecture, which is a simplified U-Net~\cite{liu2024residual}. 
In practice, we set the diffusion step to $T=4$. 
The overall framework is optimized with the default hyperparameter and training protocol of each dehazing backbone (e.g., learning rate, number of epochs, batch size, and optimizer) to ensure fair comparisons.


\noindent\textbf{Dehazing Models and Datasets.}
We adopt four state-of-the-art image dehazing models, including FocalNet~\cite{cui2023focal}, ConvIR~\cite{cui2024revitalizing}, SGDN~\cite{fang2025guided}, and RIDCP~\cite{wu2023ridcp} to validate the effectiveness of HNDiff. 
Following prior studies, we conduct experiments on two widely used synthetic dataset, SOTS-Indoor and SOTS-Outdoor~\cite{li2018benchmarking}, and five real-world benchmarks: NH-HAZE~\cite{ancuti2021ntire}, O-HAZE~\cite{ancuti2018haze}, Dense-HAZE~\cite{ancuti2019dense}, RW\textsuperscript{2}AH~\cite{fang2025guided}, and RTTS~\cite{li2018benchmarking}. 
The SOTS-Indoor dataset consists of 13,990 training pairs and 500 testing pairs.
The SOTS-Outdoor dataset consists of 313,950 training pairs and 500 testing pairs.
Both NH-HAZE and Dense-HAZE provide 50 training pairs and 5 testing pairs. 
O-HAZE offers 40 training pairs and 5 testing pairs. 
RW\textsuperscript{2}AH includes 1,406 training pairs and 352 testing pairs.
RTTS contains 4,322 hazy images and is used exclusively for testing.

\subsection{Experimental Results}
\begin{table*}[t]
\centering
\caption{Quantitative results on six benchmark datasets. Values in parentheses represent the improvements of HNDiff over the corresponding baselines.}
\label{tab:dehaze_main}
\resizebox{\linewidth}{!}{
\begin{tabular}{l *{6}{cc}}
\toprule
\multirow{2}{*}{Model}
& \multicolumn{2}{c}{NH-HAZE} & \multicolumn{2}{c}{O-HAZE} & \multicolumn{2}{c}{Dense-HAZE}
& \multicolumn{2}{c}{RW\textsuperscript{2}AH} & \multicolumn{2}{c}{SOTS-Indoor} & \multicolumn{2}{c}{SOTS-Outdoor} \\
\cmidrule(lr){2-3}\cmidrule(lr){4-5}\cmidrule(lr){6-7}\cmidrule(lr){8-9}\cmidrule(lr){10-11}\cmidrule(lr){12-13}
& PSNR & SSIM & PSNR & SSIM & PSNR & SSIM & PSNR & SSIM & PSNR & SSIM & PSNR & SSIM \\
\midrule
MSBDN\cite{dong2020multi}            & 17.97 & 0.659 & 24.36 & 0.749 & 15.13 & 0.555 & 21.51 & 0.595 & 33.67 & 0.985 & 33.48 & 0.982 \\
FFA-Net\cite{qin2020ffa}          & 18.13 & 0.647 & 22.12 & 0.770 & 15.70 & 0.549 & 18.73 & 0.556 & 36.39 & 0.989 & 33.57 & 0.984 \\
Dehamer\cite{guo2022image}          & 20.66 & 0.684 & 25.11 & 0.777 & 16.62 & 0.560 & 20.84 & 0.581 & 36.63 & 0.988 & 35.18 & 0.986 \\
MB-Taylor\cite{qiu2023mb}  & 20.43 & 0.688 & 25.05 & 0.788 & 16.66 & 0.560 & 21.37 & 0.608 & 40.71 & 0.992 & 37.42 & 0.989 \\
FocalNet\cite{cui2023focal}         & 20.36 & 0.696 & 25.46 & 0.791 & 16.95 & 0.597 & 21.93 & 0.635 & 40.82 & 0.992 & 37.71 & 0.995 \\
ConvIR\cite{cui2024revitalizing}           & 20.65 & 0.692 & 25.25 & 0.784 & 16.86 & 0.600 & 21.99 & 0.640 & 41.53 & 0.994 & 37.95 & 0.994 \\
SGDN\cite{fang2025guided}             & 20.13 & 0.680 & 24.59 & 0.778 & 16.60 & 0.571 & 22.24 & 0.631 & 41.01 & 0.992 & 36.22 & 0.986 \\
\midrule
FocalNet+HNDiff & \makecell{\textbf{20.89} \\ {\scriptsize \textbf{(+0.53)}}} & \makecell{\textbf{0.697} \\ {\scriptsize \textbf{(+0.001)}}} &
                    \makecell{\textbf{26.32} \\ {\scriptsize \textbf{(+0.86)}}} & \makecell{\textbf{0.801} \\ {\scriptsize \textbf{(+0.010)}}} &
                    \makecell{\textbf{17.29} \\ {\scriptsize \textbf{(+0.34)}}} & \makecell{\textbf{0.599} \\ {\scriptsize \textbf{(+0.002)}}} &
                    \makecell{\textbf{22.29} \\ {\scriptsize \textbf{(+0.36)}}} & \makecell{\textbf{0.647} \\ {\scriptsize \textbf{(+0.012)}}} &
                    \makecell{\textbf{41.19} \\ {\scriptsize \textbf{(+0.37)}}} & \makecell{\textbf{0.994} \\ {\scriptsize \textbf{(+0.002)}}} &
                    \makecell{\textbf{38.10} \\ {\scriptsize \textbf{(+0.39)}}} & \makecell{\textbf{0.996} \\ {\scriptsize \textbf{(+0.001)}}} \\
ConvIR+HNDiff   & \makecell{\textbf{21.23} \\ {\scriptsize \textbf{(+0.58)}}} & \makecell{\textbf{0.701} \\ {\scriptsize \textbf{(+0.009)}}} &
                    \makecell{\textbf{26.20} \\ {\scriptsize \textbf{(+0.95)}}} & \makecell{\textbf{0.799} \\ {\scriptsize \textbf{(+0.015)}}} &
                    \makecell{\textbf{17.18} \\ {\scriptsize \textbf{(+0.32)}}} & \makecell{\textbf{0.623} \\ {\scriptsize \textbf{(+0.023)}}} &
                    \makecell{\textbf{22.25} \\ {\scriptsize \textbf{(+0.26)}}} & \makecell{\textbf{0.646} \\ {\scriptsize \textbf{(+0.006)}}} &
                    \makecell{\textbf{42.10} \\ {\scriptsize \textbf{(+0.57)}}} & \makecell{\textbf{0.995} \\ {\scriptsize \textbf{(+0.001)}}} &
                    \makecell{\textbf{38.83} \\ {\scriptsize \textbf{(+0.88)}}} & \makecell{\textbf{0.995} \\ {\scriptsize \textbf{(+0.001)}}} \\
SGDN+HNDiff     & \makecell{\textbf{20.64} \\ {\scriptsize \textbf{(+0.51)}}} & \makecell{\textbf{0.686} \\ {\scriptsize \textbf{(+0.006)}}} &
                    \makecell{\textbf{25.40} \\ {\scriptsize \textbf{(+0.81)}}} & \makecell{\textbf{0.782} \\ {\scriptsize \textbf{(+0.004)}}} &
                    \makecell{\textbf{17.17} \\ {\scriptsize \textbf{(+0.57)}}} & \makecell{\textbf{0.611} \\ {\scriptsize \textbf{(+0.040)}}} &
                    \makecell{\textbf{22.81} \\ {\scriptsize \textbf{(+0.57)}}} & \makecell{\textbf{0.653} \\ {\scriptsize \textbf{(+0.022)}}} &
                    \makecell{\textbf{41.47} \\ {\scriptsize \textbf{(+0.46)}}} & \makecell{\textbf{0.995} \\ {\scriptsize \textbf{(+0.003)}}} &
                    \makecell{\textbf{37.10} \\ {\scriptsize \textbf{(+0.88)}}} & \makecell{\textbf{0.991} \\ {\scriptsize \textbf{(+0.005)}}} \\
\midrule
Avg Gains &
\multicolumn{1}{c}{\textbf{+0.54}} & \multicolumn{1}{c}{\textbf{+0.005}} &
\multicolumn{1}{c}{\textbf{+0.87}} & \multicolumn{1}{c}{\textbf{+0.010}} &
\multicolumn{1}{c}{\textbf{+0.41}} & \multicolumn{1}{c}{\textbf{+0.022}} &
\multicolumn{1}{c}{\textbf{+0.40}} & \multicolumn{1}{c}{\textbf{+0.013}} &
\multicolumn{1}{c}{\textbf{+0.47}} & \multicolumn{1}{c}{\textbf{+0.002}} &
\multicolumn{1}{c}{\textbf{+0.72}} & \multicolumn{1}{c}{\textbf{+0.002}} \\
\bottomrule
\end{tabular}
}
\end{table*}

\begin{figure*}[t!]
  \centering
  \includegraphics[width=1.0\textwidth]{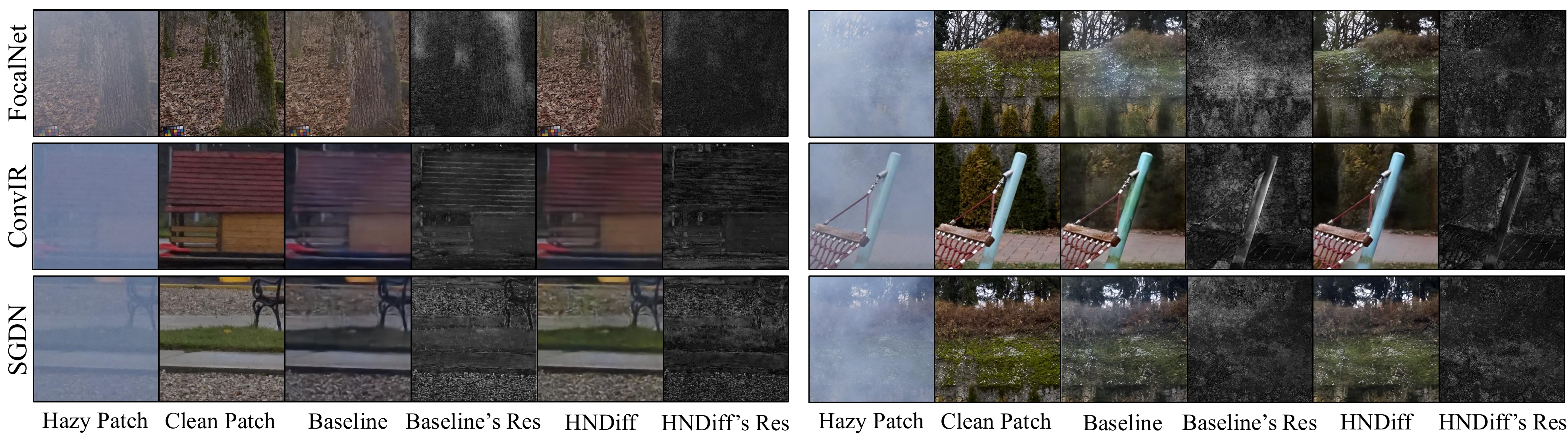} %
  \caption{Qualitative results on the O-HAZE (left) and NH-HAZE (right) datasets. “Res” denotes residual maps between outputs and ground truth, where darker intensities indicate smaller errors.}
  \label{fig:viz2.pdf}
\end{figure*}

\noindent\textbf{Quantitative Results.}
\cref{tab:dehaze_main} and \cref{tab:rtts_result} compare the dehazing performance of state-of-the-art methods and their HNDiff-enhanced versions, where the values in parentheses indicate the improvements made by HNDiff over the respective dehazing baselines. 
The results clearly demonstrate that HNDiff consistently boosts each baseline and outperforms existing state-of-the-art methods. 
Specifically, as shown in~\cref{tab:dehaze_main}, HNDiff yields average PSNR improvements of +0.54, +0.87, +0.41, +0.40, +0.47, and +0.72 dB on the NH-HAZE, O-HAZE, Dense-HAZE, RW\textsuperscript{2}AH, SOTS-Indoor, and SOTS-Outdoor test sets, respectively.
Additionally, HNDiff achieves average PSNR improvements of +0.48, +0.59, and +0.63 on baselines FocalNet, ConvIR, and SGDN, respectively. 
Overall, HNDiff achieves an average gain of +0.57 dB in PSNR and +0.009 in SSIM across all datasets and baselines, highlighting its strong generalization ability and effectiveness as a prior generation network for image dehazing.
In \cref{tab:rtts_result}, the HNDiff-enhanced RIDCP achieves the best performance among the compared methods on the real-world RTTS benchmark, attaining 0.417 FADE, 5.08 NIMA, and 16.09 BRISQUE, which indicates strong generalization and improved perceptual quality under real-world haze.
%

%

\begin{figure*}[t!]
  \centering
  \includegraphics[width=1.0\textwidth]{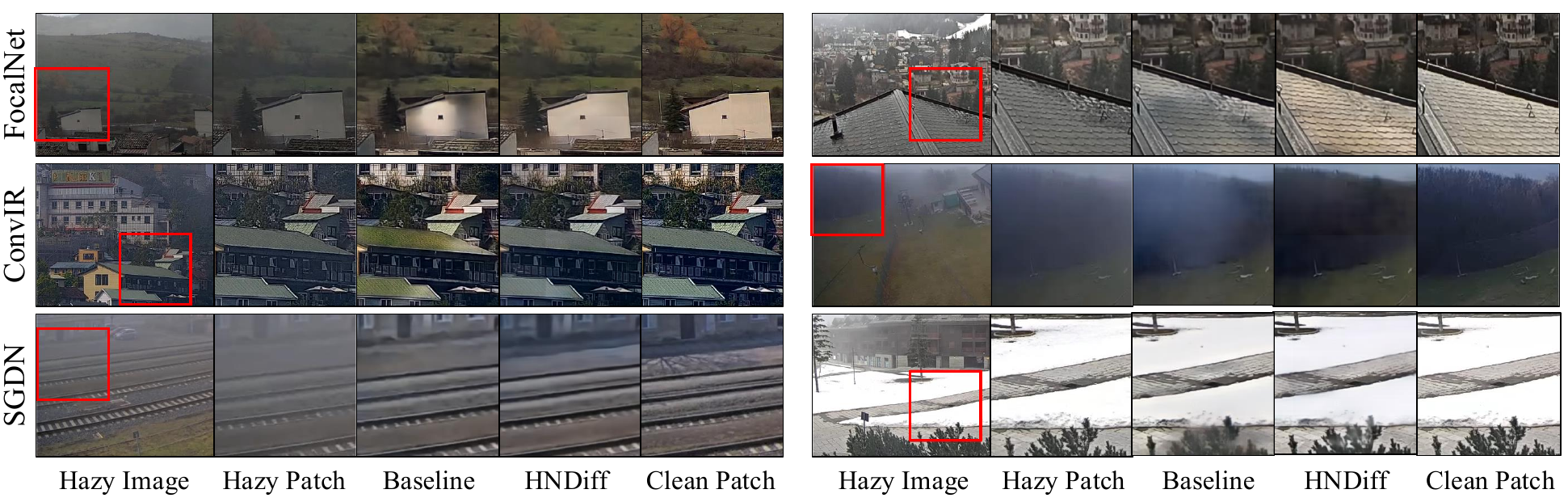} %
  \caption{Qualitative results on the RW\textsuperscript{2}AH dataset.}
  \label{fig:viz1.pdf}
\end{figure*}

\begin{figure*}[t!]
  \centering
  \includegraphics[width=1.0\textwidth]{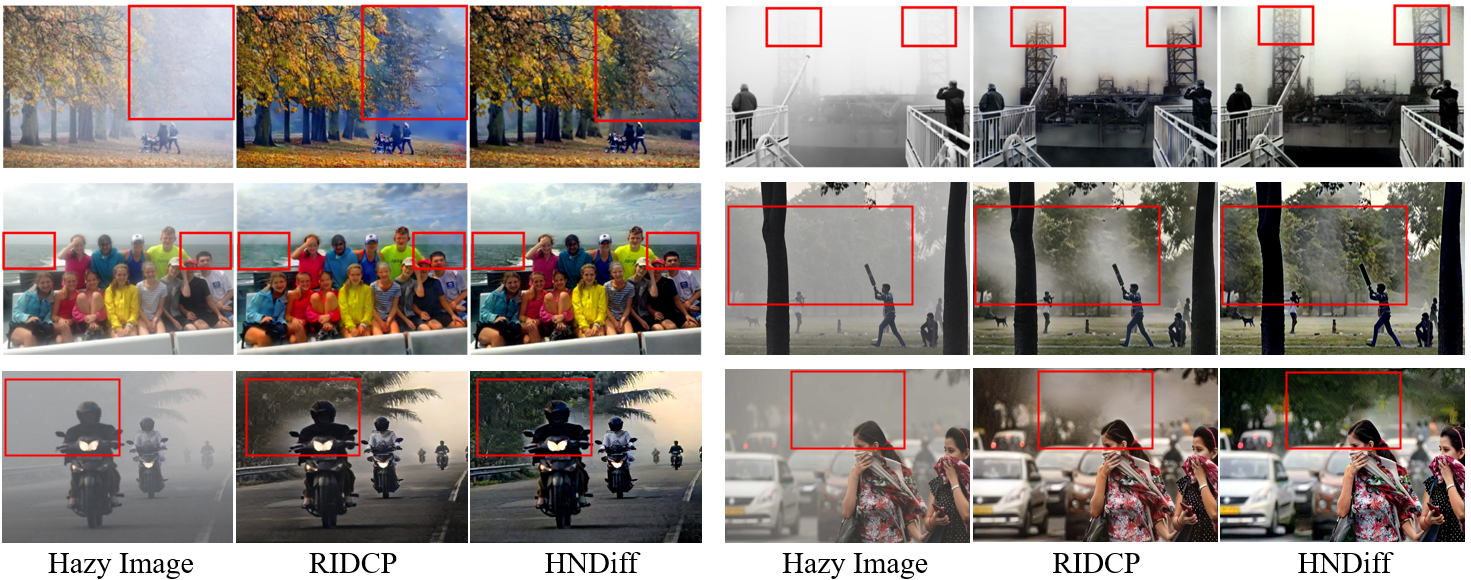}
  \caption{Qualitative results on the RTTS dataset.}
  \label{fig:rtts_viz}
\end{figure*}

\noindent\textbf{Qualitative Results.}
We present qualitative comparisons between four baseline models and their HNDiff-enhanced counterparts.
~\cref{fig:viz2.pdf} shows the results on the NH-HAZE and O-HAZE test sets, including an additional ``Res" column for better comparison. ~\cref{fig:viz1.pdf} presents the results on the RW\textsuperscript{2}AH test set. ~\cref{fig:rtts_viz} provides results on the RTTS benchmark.
The residual maps (``Res") are obtained by subtracting the ground truth from the model outputs, where lower intensities indicate smaller errors and thus higher reconstruction quality.
As shown, HNDiff consistently produces cleaner and more visually compelling dehazed images compared to the baselines. 
By integrating HNDiff into the latent space of the dehazing networks, we exploit its capacity to model rich and realistic image priors while preserving fidelity to the underlying clean image structures. 
More qualitative results are provided in the supplementary materials (Sec. 8).
%
%

\subsection{Ablation Studies}
To assess the contributions of the proposed components in HNDiff, we conduct a series of ablation studies using FocalNet as the baseline dehazing model. 
Specifically, we evaluate the effectiveness of each component, compare the prior generator with different diffusion mechanisms, analyze computational overhead and performance gain, examine HNDiff against the baseline under comparable parameter counts, inspect the haze residual modeling in latent space, compare the dehazing results of applying HNDiff in image space and latent space, and analyze the effect of varying the number of diffusion steps.
All experiments are conducted using the default training configuration of FocalNet and evaluated on the NH-HAZE test set.

\begin{table}[t]
\centering

\begin{minipage}[t]{0.42\linewidth}
\centering
\caption{Quantitative comparison on RTTS.}
\label{tab:rtts_result}
\resizebox{\linewidth}{!}{%
\begin{tabular}{l|ccc}
  \toprule
  Method & FADE$\downarrow$ & NIMA$\uparrow$ & BRISQUE$\downarrow$ \\
  \midrule
  Hazy image & 2.484 & 4.33 & 37.01 \\
  MSBDN\cite{dong2020multi}     & 1.363 & 4.14 & 28.74 \\
  Dehamer\cite{guo2022image}    & 1.895 & 3.87 & 33.87 \\
  DAD\cite{shao2020domain}      & 1.130 & 4.01 & 32.73 \\
  PSD\cite{chen2021psd}         & 0.920 & 4.35 & 25.24 \\
  RIDCP\cite{wu2023ridcp}       & 0.944 & 4.43 & 18.78 \\
  \midrule
  RIDCP+HNDiff & \textbf{0.417} & \textbf{5.08} & \textbf{16.09} \\
  \bottomrule
\end{tabular}%
}
\end{minipage}\hfill
\begin{minipage}[t]{0.55\linewidth}
\centering
\caption{Ablation study on the effectiveness of noise diffusion, haze diffusion, and HANS.}
\label{tab:ablation_diffusion}
\resizebox{\linewidth}{!}{%
  \begin{tabular}{l@{\hspace{6pt}}c@{\hspace{6pt}}c@{\hspace{6pt}}c@{\hspace{6pt}}c}
    \toprule
    Model & Noise Diffusion & Haze Diffusion & HANS & PSNR\\
    \midrule
    Net1 &  &  &  & 20.36 \\
    Net2 & \checkmark &  &  & 20.46 \\
    Net3 &  & \checkmark &  & 20.61 \\
    Net4 & \checkmark & \checkmark &  & 20.68 \\
    Net5 & \checkmark & \checkmark & \checkmark & \textbf{20.89} \\
    \bottomrule
  \end{tabular}%
}
\end{minipage}

\end{table}

\begin{table}[t]
\centering
\begin{minipage}[t]{0.42\linewidth}
  \centering
  \caption{Comparison of different prior generators.}
  \label{tab:prior_generator}
  \resizebox{\linewidth}{!}{%
    \begin{tabular}{l@{\hspace{8pt}}c@{\hspace{8pt}}c@{\hspace{8pt}}c}
      \toprule
      Model & Prior Generator & PSNR & SSIM \\
      \midrule
      Net1 & N/A    & 20.36 & 0.696 \\
      Net2 & U-Net  & 20.41 & 0.696 \\
      Net3 & DDPM   & 20.46 & 0.695 \\
      Net4 & RDDM   & 20.43 & 0.690 \\
      Net5 & HNDiff & \textbf{20.89} & \textbf{0.697} \\
      \bottomrule
    \end{tabular}%
  }
\end{minipage}\hfill
\begin{minipage}[t]{0.54\linewidth}
  \centering
  \caption{Computational overhead, peak training VRAM, and PSNR gains of HNDiff.}
  \label{tab:complexity}
  \setlength{\tabcolsep}{3pt}
  \renewcommand{\arraystretch}{0.95}
  \resizebox{\linewidth}{!}{%
    \begin{tabular}{lcccc}
      \toprule
      Model & Params (M) & FLOPs (G) & VRAM (M) & Gains (dB) \\
      \midrule
      FocalNet & 3.74 & 30.53 & 2854 & -- \\
      + HNDiff & 7.82 & 36.38 & 3932 & \textbf{+0.48} \\
      \midrule
      ConvIR & 5.51 & 41.96 & 4506 & -- \\
      + HNDiff & 9.59 & 45.39 & 5550 & \textbf{+0.59} \\
      \midrule
      SGDN & 11.09 & 52.95 & 22980 & -- \\
      + HNDiff & 15.30 & 56.38 & 23914 & \textbf{+0.63} \\
      \bottomrule
    \end{tabular}%
  }
\end{minipage}
\end{table}

\noindent\textbf{Effectiveness of Each Component.}
Our ablation study, detailed in~\cref{tab:ablation_diffusion}, evaluates the contribution of each component in HNDiff. 
\textit{Net1} denotes the baseline dehazing model. 
\textit{Net2} represents a conventional DDPM-based variant that employs only noise diffusion, while \textit{Net3} is a variant that incorporates only haze diffusion and omits noise diffusion. 
\textit{Net4} adopts both noise and haze diffusion but excludes the haze-aware noise scheduler (HANS). 
Finally, \textit{Net5} is the complete HNDiff design. 
The results show that both \textit{Net2} and \textit{Net3} surpass the baseline, demonstrating the individual benefits of noise and haze diffusion. 
Moreover, \textit{Net5} achieves the best performance, indicating that the joint integration of both diffusion processes together with HANS provides complementary gains. 
These findings highlight the importance of incorporating haze-aware design in order to enhance dehazing effectiveness.

\noindent\textbf{Comparison of Prior Generators with Different Diffusion Mechanisms.}
\cref{tab:prior_generator} evaluates the baseline dehazing model augmented with different prior generation methods, including U-Net, DDPM~\cite{ho2020denoising}, RDDM~\cite{liu2024residual}, and our proposed HNDiff. 
\textit{Net1} denotes the baseline model without a prior generator. 
\textit{Net2} employs a U-Net to generate priors directly, without any diffusion process. 
\textit{Net3}, \textit{Net4}, and \textit{Net5} are the dehazing models enhanced with priors generated by DDPM, RDDM, and HNDiff, respectively. 
Although integrating the standard diffusion process (\textit{Net3}) or the residual diffusion process (\textit{Net4}) improves performance over the baseline, the gain is just comparable to that of \textit{Net2}, which uses a U-Net without diffusion. 
In contrast, HNDiff explicitly embeds the atmospheric scattering model into the diffusion process, yielding consistent and superior improvements compared to both standard and residual diffusion mechanisms. We present the dehazed results of different diffusion mechanisms in the supplementary materials.

\noindent\textbf{Analysis of Computational Overhead and Performance Gain.}
We evaluate the computational overhead using cropped $256 \times 256$ patches on a single NVIDIA RTX 4090 GPU with a batch size of $4$. The reported VRAM denotes the memory required during training under this setting. As shown in~\cref{tab:complexity}, the PSNR gain is computed by averaging the improvements over the six datasets reported in~\cref{tab:dehaze_main}. HNDiff consistently improves FocalNet, ConvIR, and SGDN by +0.48 dB, +0.59 dB, and +0.63 dB, respectively. Although our method introduces additional parameters and computational complexity, the achieved PSNR gains are  substantial, demonstrating a favorable trade-off between computational cost and restoration performance.

\begin{figure*}[t!]
  \centering
  \includegraphics[width=1.0\textwidth]{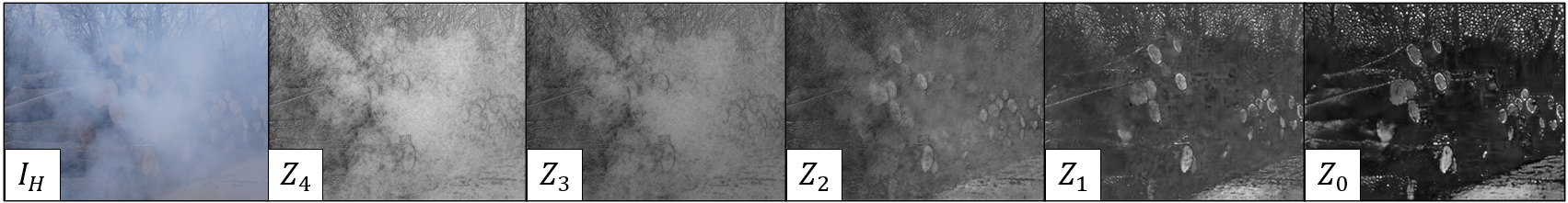} %
  \caption{Visualization of latent representations across reverse diffusion steps.}
  \label{fig:viz3.pdf}
\end{figure*}


\begin{table}[t]
\centering

\begin{minipage}[t]{0.48\linewidth}
  \centering
  \caption{Comparison of applying HNDiff in image space and latent space.}
  \label{tab:rw2ah_image_vs_latent}
  \resizebox{\linewidth}{!}{%
    \begin{tabular}{lccc}
      \toprule
      Metric & FocalNet & HNDiff (Image) & HNDiff (Latent) \\
      \midrule
      PSNR (dB) & 21.18  & 21.37  & \textbf{21.52} \\
      SSIM      & 0.5970 & 0.6166 & \textbf{0.6254} \\
      FLOPs (G) & 30.53  & 65.59  & 36.38 \\
      \bottomrule
    \end{tabular}%
  }
\end{minipage}\hfill
\begin{minipage}[t]{0.48\linewidth}
  \centering
  \caption{Comparison between HNDiff and baseline variants with comparable parameter counts.}
  \label{tab:enlarged_variants}
  \resizebox{\linewidth}{!}{%
    \begin{tabular}{lcccc}
      \toprule
      & FocalNet & FocalNet\textsuperscript{+} & FocalNet\textsuperscript{*} & HNDiff \\
      \midrule
      Params (M) & 3.74  & 8.40  & 8.28  & 7.82 \\
      FLOPs (G)  & 30.53 & 68.54 & 64.05 & 36.38 \\
      PSNR (dB)  & 20.36 & 20.37 & 20.51 & \textbf{20.89} \\
      \bottomrule
    \end{tabular}%
  }
\end{minipage}

\end{table}

\begin{figure*}[t!]
  \centering
  \includegraphics[width=1.0\textwidth]{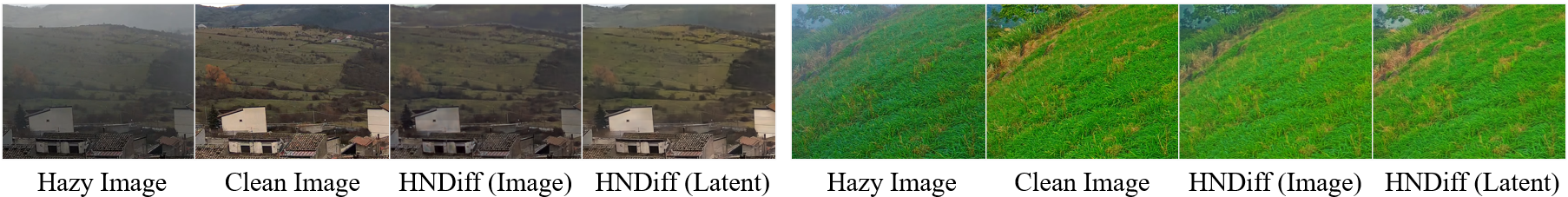} %
  \caption{Qualitative results of applying HNDiff in image space and latent space.}
  \label{fig:HNDiff_image.pdf}
\end{figure*}

\noindent\textbf{Comparison Under Comparable Parameter Counts.}
To ensure a fair comparison under similar parameter budgets, we evaluate HNDiff against enlarged baseline variants, as reported in~\cref{tab:enlarged_variants}.
Following the model-scaling study~\cite{tan2019efficientnet}, we design these variants by scaling network width and depth.
Specifically, \textit{FocalNet\textsuperscript{+}} increases the base channel size from 32 to 48, while \textit{FocalNet\textsuperscript{*}} expands the number of residual blocks from 4 to 10.
Although both variants substantially increase model complexity in terms of parameters and FLOPs, they yield only marginal PSNR gains over the baseline FocalNet. 
In contrast, HNDiff achieves the best performance of 20.89 dB in PSNR with a lower parameter count (7.82M) and significantly reduced FLOPs (36.38G). 
These results demonstrate that integrating the proposed diffusion prior is more effective than simply scaling the network capacity.



\noindent\textbf{Analysis of Haze Residual Modeling in Latent Space.}
To verify that HNDiff models haze formation in latent space, we analyze diffusion prior outputs across reverse steps.
Although the model is trained with $T=4$ steps, we examine intermediate latent representations by performing $t \in [0,1,2,3,4]$ reverse steps starting from the fully hazy latent $Z_4$, resulting in the sequence $[Z_4, Z_3, \ldots, Z_0]$. 
For visualization, we compute the channel-wise mean of each latent and downsample $I_H$ to the same spatial resolution only for visualization.
As shown in~\cref{fig:viz3.pdf}, the representations progressively transition from hazy ($Z_4$) to clean ($Z_0$), confirming that HNDiff captures a progressive hazy-to-clean structure in the latent space and enables interpretable modeling.

\noindent\textbf{Comparison of Applying HNDiff in Image Space and Latent Space.}
\cref{tab:rw2ah_image_vs_latent} compares an image-space variant, \emph{HNDiff (Image)}, which applies our haze-noise diffusion directly to RGB images, and a latent-space variant, \emph{HNDiff (Latent)}, which operates in the latent space of a FocalNet on the real-world RW\textsuperscript{2}AH dataset. The two variants use the same U-Net as the haze/noise estimator and both improve over the baseline, while \emph{HNDiff (Latent)} achieves the best performance with only minimal additional FLOPs overhead, confirming its advantage in content fidelity. \cref{fig:HNDiff_image.pdf} further shows that \emph{HNDiff (Latent)} produces results closer to the ground truth, supporting our choice of the latent formulation in the main experiments.

\begin{figure*}[t!]
  \centering
  \includegraphics[width=1.0\textwidth]{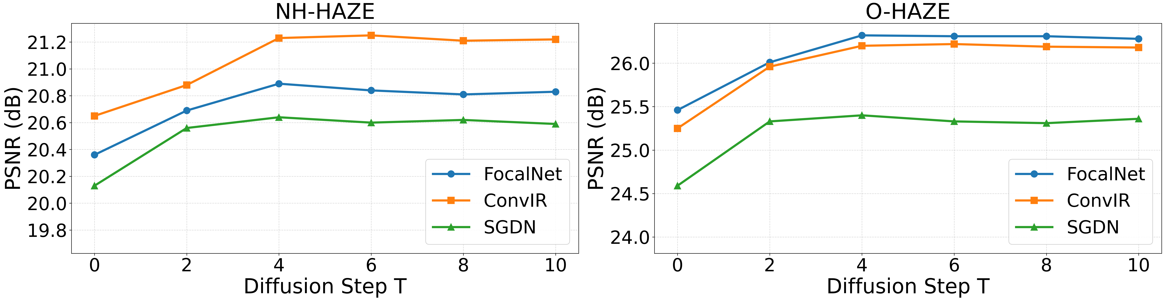} %
  \caption{Effect of the diffusion step $T$ across three dehazing backbones on NH-HAZE and O-HAZE.}
  \label{fig:diffusion_step}
\end{figure*}

\noindent\textbf{Analysis of Diffusion Step Setting.}
\cref{fig:diffusion_step} analyzes the impact of varying diffusion steps $T$ across three restoration backbones on the NH-HAZE and O-HAZE datasets.
Without diffusion guidance ($T=0$), performance is limited for all backbones.
Increasing $T$ improves the results, with FocalNet and SGDN peaking at $T=4$, and ConvIR peaking at $T=6$.
Further increasing $T$ brings no additional gains.
These results show that our method achieves strong performance with only a few diffusion steps.

\section{Limitations}
HNDiff is tailored to the Atmospheric Scattering Model and thus cannot be directly applied to other degradations such as motion blur, raindrops, or low-light conditions. Extending it to these scenarios requires integrating degradation-specific priors (e.g., object motion, rain masks, exposure time), which we leave as future work.
\section{Conclusion}
We propose Haze-Noise Diffusion (HNDiff), a novel diffusion-based framework for image dehazing.
HNDiff integrates the atmospheric scattering model into the diffusion framework, jointly performing haze diffusion and noise diffusion to account for the physical properties of haze formation.
In the forward process, HNDiff progressively degrades a clean image by introducing both haze and noise through a haze-noise diffusion, with a haze-aware noise scheduler that adaptively adjusts noise levels according to haze density. 
In the reverse process, HNDiff restores the image by removing both haze and noise through its dehazing-denoising process. 
To enhance the existing dehazing methods, we incorporate HNDiff into their latent spaces as a prior generator, seamlessly integrating the learned prior into each encoder/decoder block via our proposed Feature Gating Module to generate higher-quality dehazed results.  
Extensive experimental results have demonstrated that our method effectively improves the performance of four state-of-the-art dehazing models across seven dehazing datasets.

%
%
%
%
\section*{Acknowledgments}
This work was supported in part by the National Science and Technology Council (NSTC) under grants 114-2221-E-A49-038-MY3, 115-2634-F-A49-011, 113-2221-E-004-001-MY3, 114-2221-E-007-065-MY3 and by the NVIDIA Taiwan AI Research \& Development Center (TRDC). This work was supported by H100 GPU computing resources donated by Wistron.

%
%
\bibliographystyle{splncs04}
\bibliography{main}
\end{document}